\documentclass[11pt]{article}
\usepackage{eacl2017}
\usepackage{times}
\usepackage{url}
\usepackage{latexsym}
\usepackage{graphics}
\usepackage{longtable}
\usepackage{adjustbox}
\usepackage[caption=false]{subfig}
\usepackage{float}
\usepackage{multirow}
\usepackage{arydshln}

\eaclfinalcopy 

\title{Evaluating Word Embedding Hyper-Parameters for \\Similarity and Analogy Tasks}

\author{
	Maryam Fanaeepour{$~^{1,2}$} \, Adam Makarucha{$~^{2}$} \, Jey Han Lau{$~^{2}$}\\
	  {\small \hspace{4em} maryam@cs.duke.edu \hspace{3.5em} adamjm@au1.ibm.com \, \hspace{1.5em}jeyhan.lau@au1.ibm.com} \\
	$^{1}$\, Department of Computer Science, Duke University, USA\\
	$^{2}$\, IBM Research, Melbourne, Australia
}

\date{}

\begin{document}
\maketitle

	

\begin{abstract}
\noindent The versatility of word embeddings for various applications is 
attracting researchers from various fields. However, the impact of 
hyper-parameters when training embedding model is often poorly 
understood. How much do hyper-parameters such as vector dimensions and 
corpus size affect the quality of embeddings, and how do these results 
translate to downstream applications?  Using standard embedding 
evaluation metrics and datasets, we conduct a study to empirically 
measure the impact of these hyper-parameters.

\end{abstract}


\section{Introduction}\label{sec:intro}
Vector representations of words have been widely utilised in various 
applications, from natural language processing tasks 
\cite{Bengio+:2003,Zhang+:2014a,Li+:2014a} to object recognition 
\cite{Frome+:2013}.  A number of methods to create these vector representations (a.k.a word 
embeddings) have been developed, such as Skip-gram Negative Sampling (a.k.a. word2vec: 
\newcite{Mikolov13}) and GloVe~\cite{PenningtonSM14}.  Swivel  
\cite{swivel16} has recently been proposed as a method to generate word embeddings and has been shown to be a 
competitive methodology. Swivel has been used as the word embedding model in this work.

Intrinsic tasks such as similarity and analogy test for syntactic or 
semantic relationship between words~\cite{MikolovSCCD13} using the raw 
vectors of words learnt by the embedding models. In this paper, we use 
these intrinsic tasks as part of our evaluation. We vary a number of   
hyper-parameter settings to measure their impact on the performance of the task.
Contrary to expectation, we find that, for example, increasing the 
corpus size has little impact on word similarity tasks.



\section{Related Work}\label{sec:related_work}
There are a number of studies on evaluating the quality of the word 
embeddings~\cite{Chiu16,Schnabel15,Linzen16,Gladkova16}.  
\newcite{Chiu16} studied the relationship between intrinsic and extrinsic 
task performance produced by word embeddings.  The authors found that 
models that performed well in intrinsic tasks do not necessarily perform 
well in downstream tasks such as sequence labelling problems, suggesting 
the  limited utility of using intrinsic tasks for evaluating word 
embeddings.

\newcite{FaruquiTRD16} suggested using task-specific evaluation for 
word-embeddings, since different types of information was captured by 
different embedding models.

\newcite{Levy+:2014} analysed skip-gram and discovered that it is 
implicitly factorising a word-context matrix, revealing its relationship 
with traditional vector generation approaches such as singular value 
decomposition (SVD). The discovery provided theoretical explanation for 
the successes of skip-gram and neural embeddings in general.

\newcite{Levy:2015b} compared several word embedding methodologies, and 
found that the strong performance of a particular embedding methodology 
could be due to system design choices and hyper-parameter settings. When 
all methodologies are standardised to using similar hyper-parameter 
settings, the authors found little performance difference between the 
different embedding methodologies.

\section{Methodology}\label{sec:methodology}
The methodology section is structured as follows, 
firstly a description of Swivel, the embedding methodology that was 
used to generate the word embeddings used for all experiments. 
Then a description of the set of hyper-parameters 
that were explored, followed by the evaluation metrics and datasets.

\subsection{Swivel}\label{sec:swivel}

We use Swivel (Submatrix-wise Vector Embedding Learner: 
\newcite{swivel16}), a method that generates low-dimensional feature 
embeddings from a feature co-occurrence matrix. It performs approximate 
factorisation of the point-wise mutual information (PMI) matrix between 
each row and column features via stochastic gradient descent. Swivel 
uses a piecewise loss with special handling for unobserved 
co-occurrences. To improve computational efficiency, it makes use of 
vectorized multiplication to process thousands of rows and columns at 
once to compute millions of predicted values. The matrix is partitioned 
into sub-matrices to parallelize the computation across many nodes, 
allowing Swivel to scale for large corpora.

In detail, Swivel represents a $m \times n$ co-occurrence matrix between 
$ m $ row and $ n $ column features, in which row feature and each 
column feature are assigned individually a $ d $ dimensional embedding 
vector.  The vectors are grouped into blocks or submatrix called 
``shard".  Training proceeds by selecting a shard (and thus, its 
corresponding row block and column block), and performing a matrix 
multiplication of the associated vectors to produce an estimate of the 
PMI values for each co-occurrence.  This is compared with the observed 
PMI, with special handling for the case where no co-occurrence was 
observed and the PMI is undefined.  Stochastic gradient descent is used 
to update the individual vectors and minimise the difference. Swivel 
uses a piecewise loss function to differentiate between observed and 
unobserved co-occurrences. Let $x_{ij}$ be the number of the times the 
focus word $i$ co-occurs with the context word $j$. The training 
objective is given as follows. If $x_{ij} > 0$ (co-occurrence is 
observed), swivel computes the weighted squared error of the difference 
between the dot product of the embeddings and the PMI of $i$ and $j$.  
In other words, the model is optimised to predict the observed PMI 
score.  For unobserved co-occurrence, i.e.\ $x_{ij} = 0$, soft hinge is 
the cost function, where a smoothed PMI is used by assuming it has an 
actual count of 1. It is crucial that the model does not overestimate 
the PMI of common words whose co-occurrence is unobserved.

\subsection{Pre-processing}
Our training data was based on English Wikipedia.\footnote{Wikipedia dump 
retrieved on June 2016 Wikipedia}.  The text was partitioning it into several sizes, presented in 
Table~\ref{table:experiment}. Gensim \cite{gensim} was used to lowercase 
and tokenise the corpus, and discard all punctuation.

\begin{table}[h!]
	\centering
	\begin{adjustbox}{width=\columnwidth}
        \begin{tabular}{rcc}
            \textbf{Data Size} & \textbf{Vocabulary Size} & 
            \textbf{Token Counts}\\
            \hline
            Small & 204,800 & 798,857 \\
            Medium & 311,296 & 1,236,639 \\
            Large  & 991,232 & 4,116,618 \\
        \end{tabular}
	\end{adjustbox}
    \caption{Training corpora statistics.}\label{table:experiment}
\end{table}


\subsection{Hyper-Parameter Settings}\label{sec:exp-param}

The following hyper-parameters were explored: window size, vector dimension 
size and corpus size.

\subsection{Evaluation Datasets and Metrics}\label{sec:evaluation}
Model performance was evaluated on two intrinsic tasks: word similarity 
and analogy. Several publicly available datasets were used, detailed in
Table~\ref{table:datasets}.

\begin{table} [h!]
        \label{table:datasets}
        \centering{
                \begin{adjustbox}{width=1\columnwidth}
            \begin{tabular}{rcl}
                \multirow{2}{*}{\textbf{Dataset}} & \textbf{Word Pairs/}
                & \multirow{2}{*}{\textbf{Reference}} \\
                & \textbf{Questions} & \\
                \hline
                                MEN & 3000 & \cite{MENbruni12}\\
                M. Turk & 287 & \cite{MTurkRadinskyAGM11} \\
                Rare Words & 2034 & \cite{RareLuongSM13} \\
                SimLex & 999 & \cite{SimLexHillRK15}\\
                \multirow{3}{*}{Word Relatedness} & \multirow{3}{*}{353}
                & \cite{WordSimFinkelstein2001};\\
                &&\cite{AgirreAHKPS09};\\
                &&\cite{ZeschMG08}\\
                \multirow{3}{*}{Word Similarity} & \multirow{3}{*}{353}
                & \cite{WordSimFinkelstein2001};\\
                &&\cite{AgirreAHKPS09};\\
                &&\cite{ZeschMG08}\\
                \multirow{2}{*}{Google} & \multirow{2}{*}{8,000} &
                \cite{Mikolov13};\\
                &&\cite{MikolovSCCD13}\\
                \multirow{2}{*}{MSR} & \multirow{2}{*}{19,544} &
                \cite{Mikolov13};\\
                &&\cite{MikolovSCCD13}\\

                        \end{tabular}
                \end{adjustbox}
        }
        \caption{Word similarity and analogy datasets.}
\end{table}

These word similarity datasets contain word pairs with human-assigned 
similarity scores.  To evaluate the word vectors, the pairs are ranked 
based on their cosine similarities, and the Spearman's rank 
correlation coefficient ($\rho$) was computed between the ranked order produced by the cosine 
similarity and that of the human ratings.

For the analogy task the system is required to infer 
word $x$, given words $a$, $b$ and $y$, where the $x$ and $y$ share the 
same relationship (e.g. \textit{city-country}) as $a$ and $b$.  
Formally, the system computes:
\[
argmax_{x} \  cos(v(x), v(a)-v(b)+v(y))
\]
where $v(w)$ is the vector of word $w$; and $cos$ is the cosine 
similarity function.

The Google dataset contains both syntactic and semantic analogies, while 
the MSR  dataset contains only syntactic analogies. In terms of 
evaluation metric, it uses accuracy, which is a ratio of the number of 
correctly answered questions to all questions.

\section{Results}

In all plots to follow, red lines denote word similarity evaluation 
performance, and blue lines denote analogy performance. For both 
evaluations, out-of-vocabulary words are discarded.

\subsection{Varying Window Sizes}

\begin{figure}[t!]
    \begin{minipage}[t]{1\columnwidth}
        \centering
        \subfloat[Dimension 20.\label{fig:size1-dim20}]{
            \includegraphics[scale = .13]{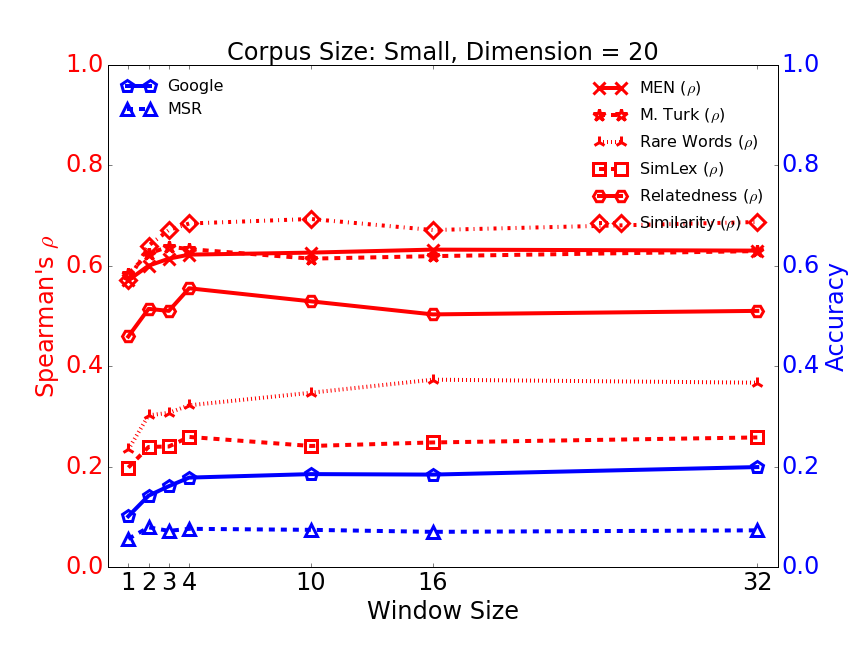}}
        \subfloat[Dimension 100.\label{fig:size1-dim100}]{
            \includegraphics[scale = .13]{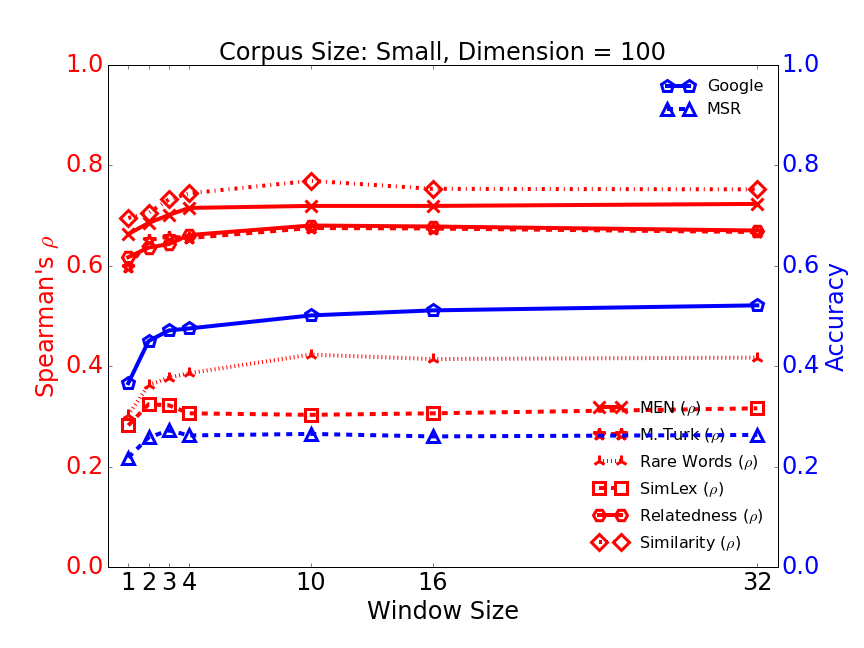}}
		
		\subfloat[Dimension 200.\label{fig:size1-dim200}]{
			\includegraphics[scale = .13]{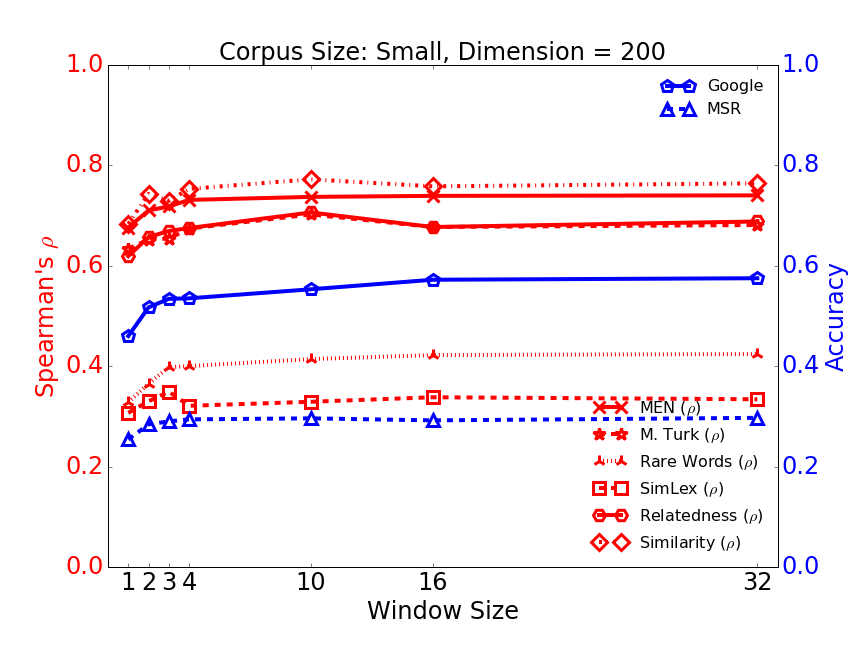}}
		\subfloat[Dimension 500.\label{fig:size1-dim500}]{
			\includegraphics[scale = .13]{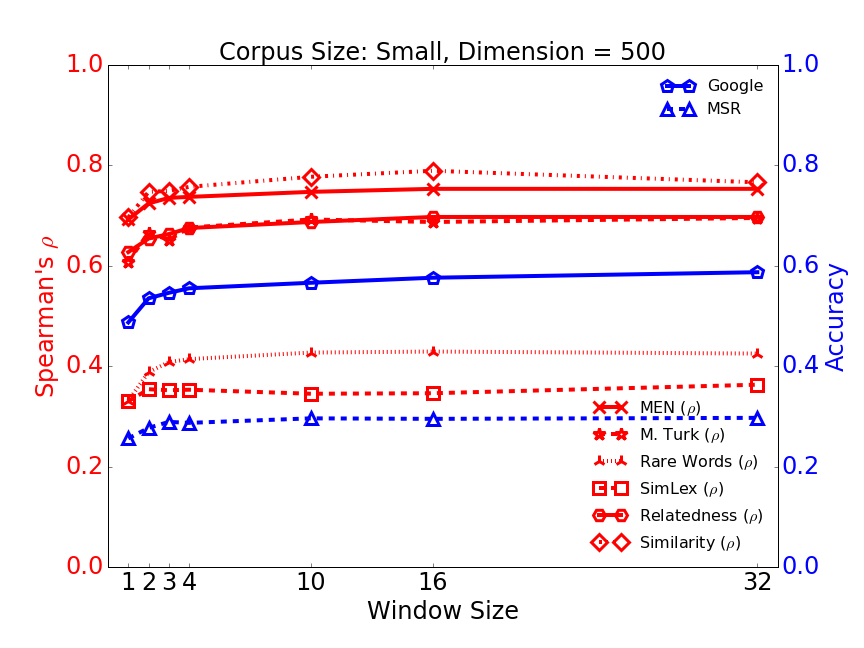}}
        \caption{Task performance with varying window sizes (corpus size 
        $=$ small).}
		\label{fig:corpus_small_wind}
	\end{minipage}
\end{figure}

\begin{figure}[t!]
	\begin{minipage}[t]{1\columnwidth}
		\centering
		\subfloat[Dimension 20.\label{fig:size3-dim20}]{
			\includegraphics[scale = .13]{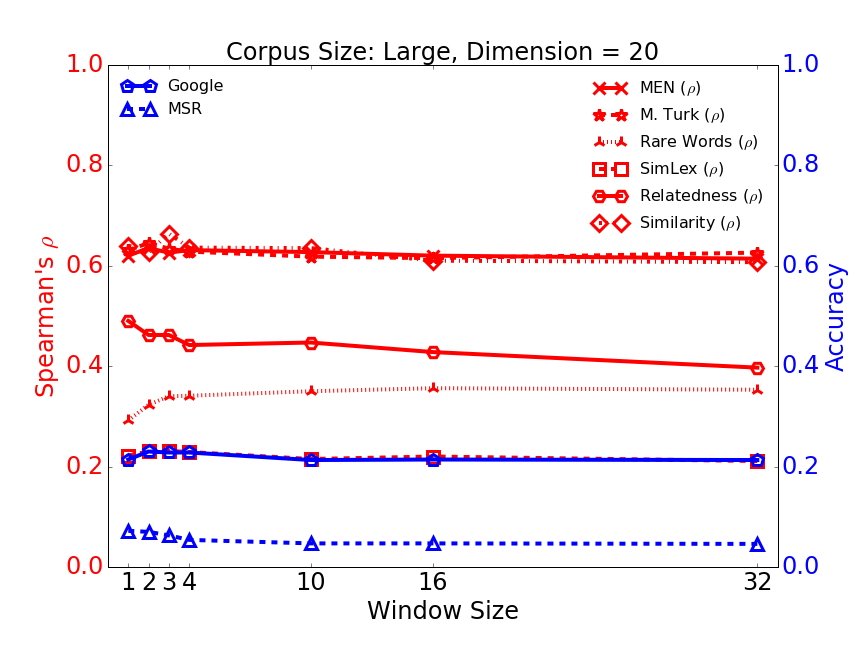}}
		\subfloat[Dimension 100.\label{fig:size3-dim100}]{
			\includegraphics[scale = .13]{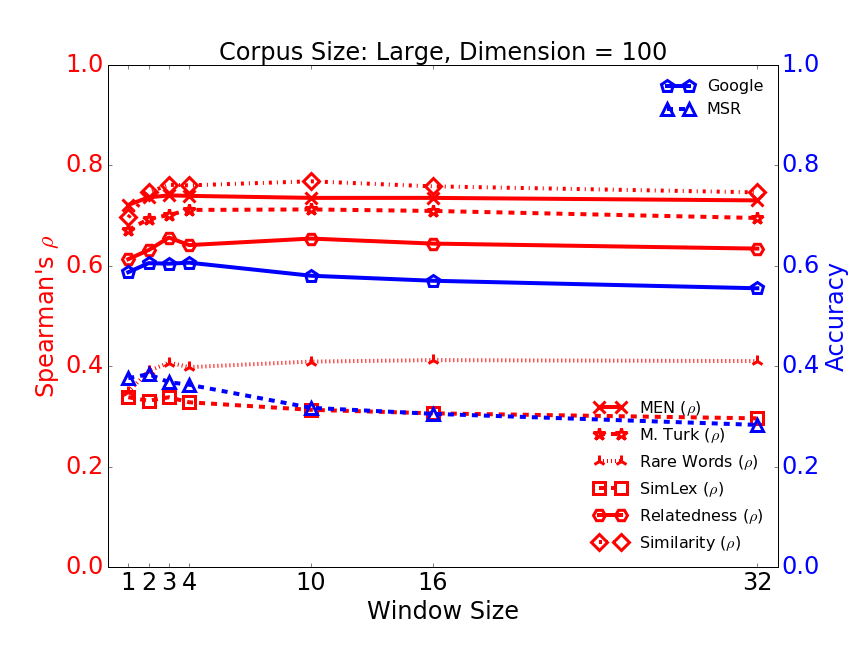}}
		
        \subfloat[Dimension 200.\label{fig:size3-dim200}]{
            \includegraphics[scale = .13]{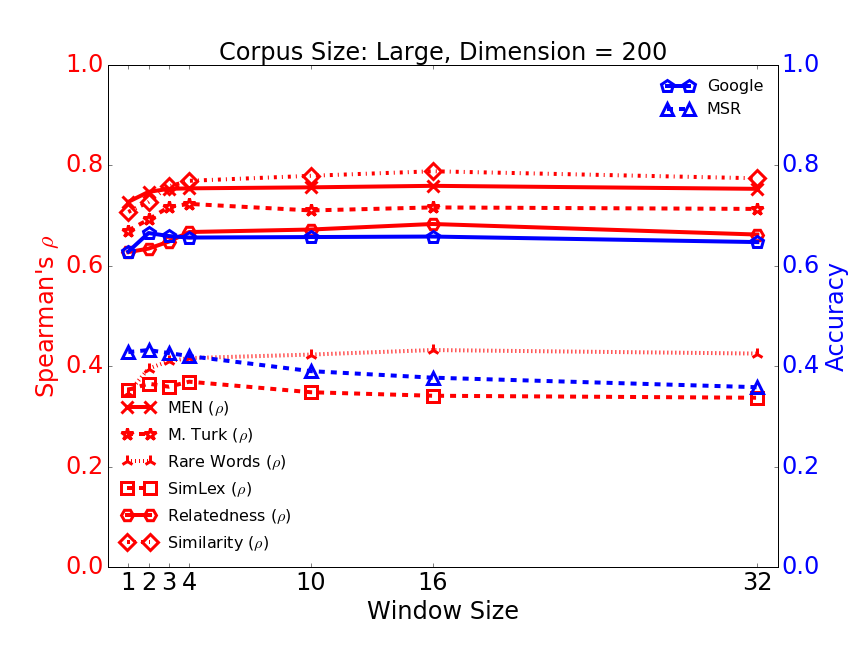}}
        \subfloat[Dimension 500.\label{fig:size3-dim500}]{
            \includegraphics[scale = .13]{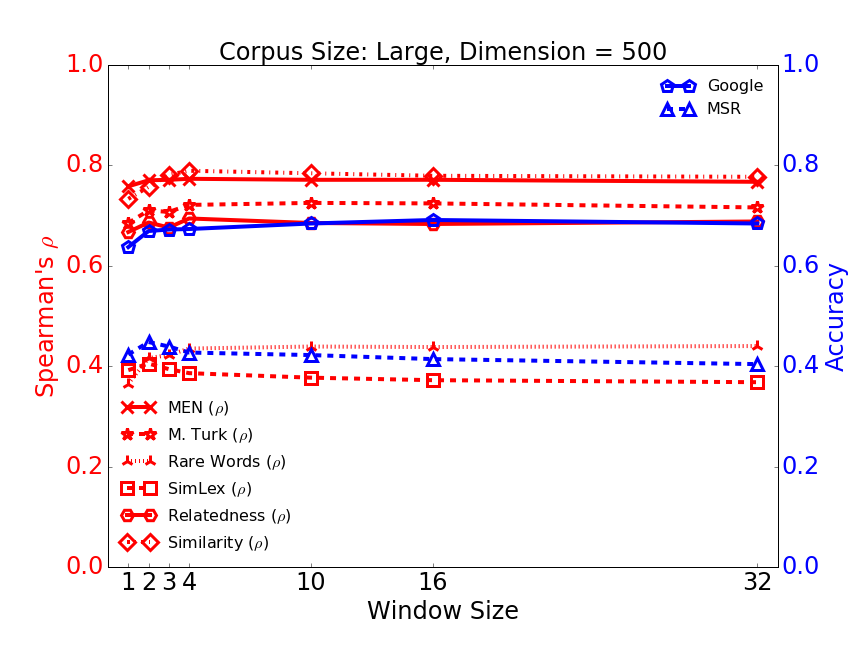}}
        \caption{Task performance with varying window sizes (corpus size 
        $=$ large).}
        \label{fig:corpus_large_wind}
    \end{minipage}
\end{figure}

The window size hyper-parameter controls the number of contextual words 
surrounding a target word. The window size varied from 
1 to 32. Results are presented in Figures~\ref{fig:corpus_small_wind} 
  (corpus size $=$ small) and \ref{fig:corpus_large_wind} (corpus size 
$=$ large)

In each figure, results for varying vector dimension 
sizes are also shown.  In general, it was found that there was little difference for both the similarity and analogy tasks when window size was greater than 4. This implies that increasing 
the window size beyond 4 does not have a significant impact on the performance in 
either task. 

\subsection{Varying Dimension Sizes}

\begin{figure}[t!]
	\begin{minipage}[t]{1\columnwidth}
		\centering
		\subfloat[Window size 2.\label{fig:size3-wind2}]{
			\includegraphics[scale = .13]{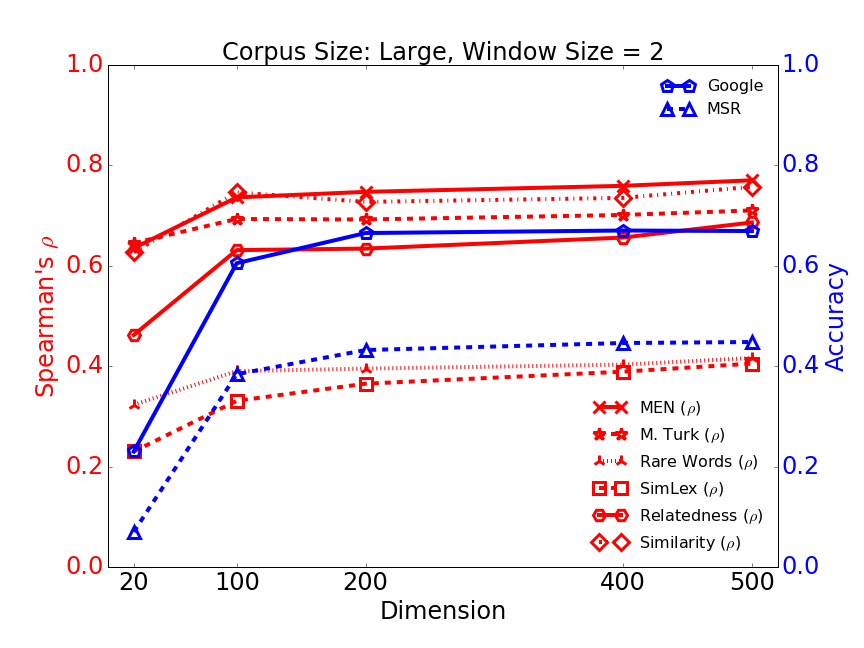}}
		\subfloat[Window size 4.\label{fig:size3-wind4}]{
			\includegraphics[scale = .13]{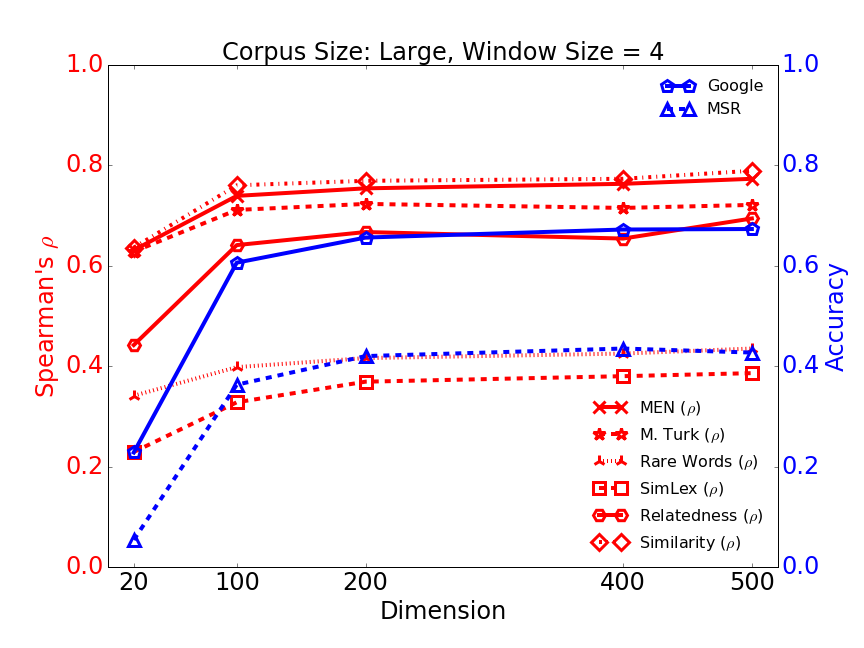}}
		
		\subfloat[Window size 10.\label{fig:size3-wind10}]{
			\includegraphics[scale = .13]{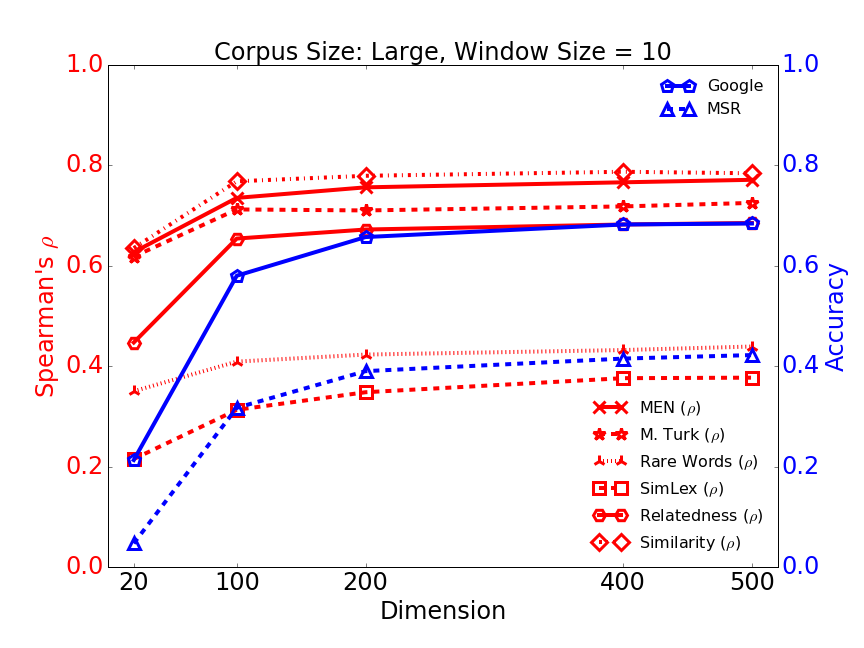}}
		\subfloat[Window size 32.\label{fig:size3-wind32}]{
			\includegraphics[scale = .13]{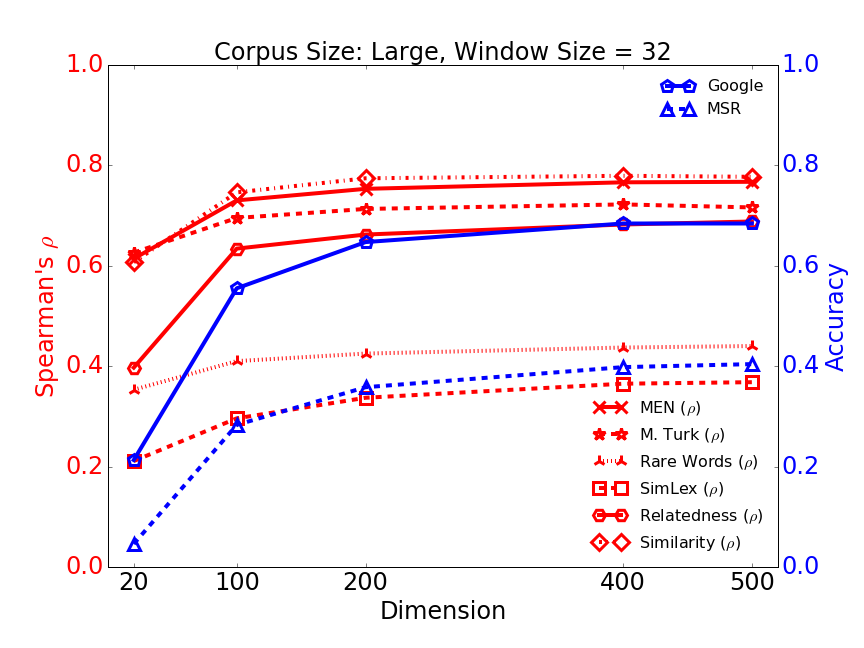}}
        \caption{Task performance with varying dimension sizes.}
		\label{fig:corpus_large_dim}
	\end{minipage}
\end{figure}

The dimension of the vectors was varied from 20 to 500; results are 
presented in Figure~\ref{fig:corpus_large_dim} (corpus size $=$ large).
Largely similar results were observed with all corpus sizes;
for brevity results using only the large corpus are presented. 
  
The performance changes very little when dimension $= 200$ was reached,
with the exception of the analogy tasks where a small improvement was observed when the window size was greater than 4. 
These finding could be useful in situations where computational resources are limited, 
as it suggests that increasing the embedding dimension (and thus the number of parameters) does not necessarily translate to improved performance.

%

\subsection{Varying Corpus/Token Sizes}

\begin{figure}[t!]
    \begin{minipage}[t]{1\columnwidth}
        \centering
        \subfloat[Window size 2.\label{fig:wind2-dim400}]{
            \includegraphics[scale = .13]{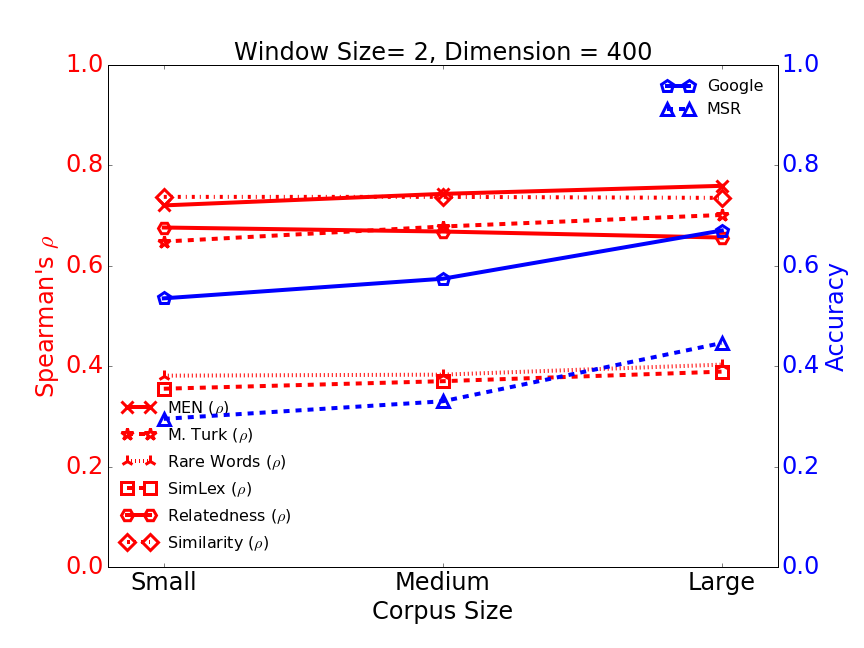}}
        \subfloat[Window size 4.\label{fig:wind4-dim400}]{
            \includegraphics[scale = .13]{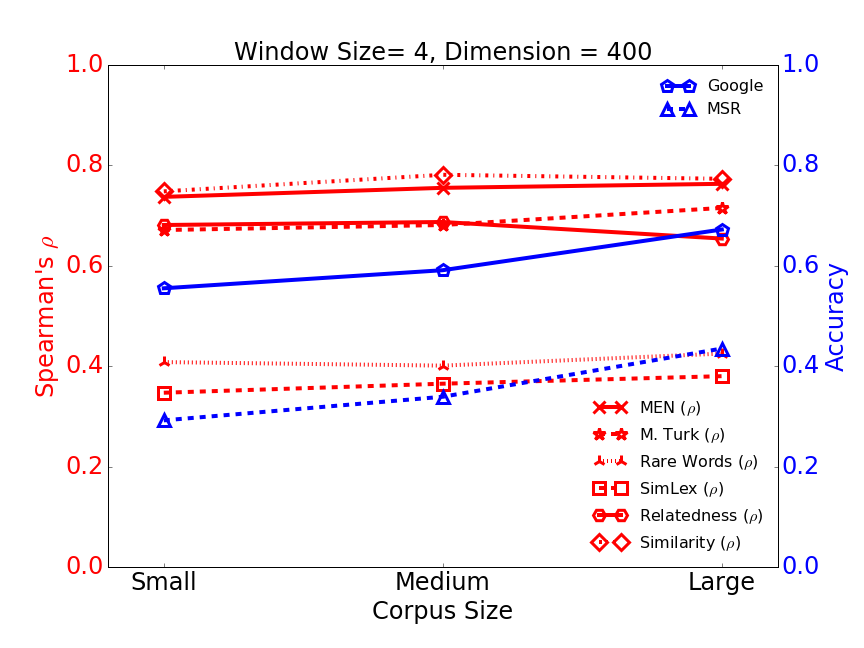}}
		
		\subfloat[Window size 10.\label{fig:wind10-dim400}]{
			\includegraphics[scale = .13]{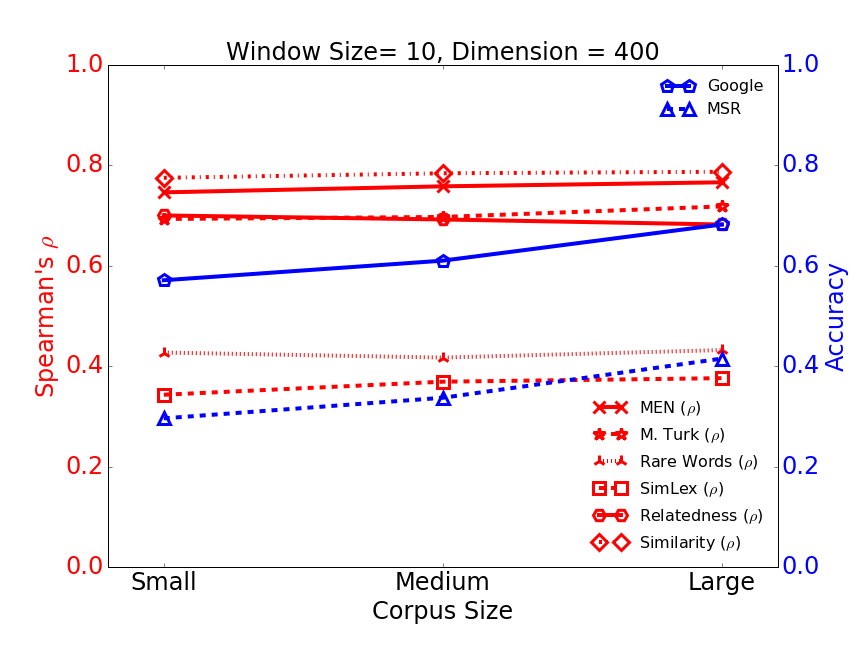}}
		\subfloat[Window size 32.\label{fig:wind32-dim400}]{
			\includegraphics[scale = .13]{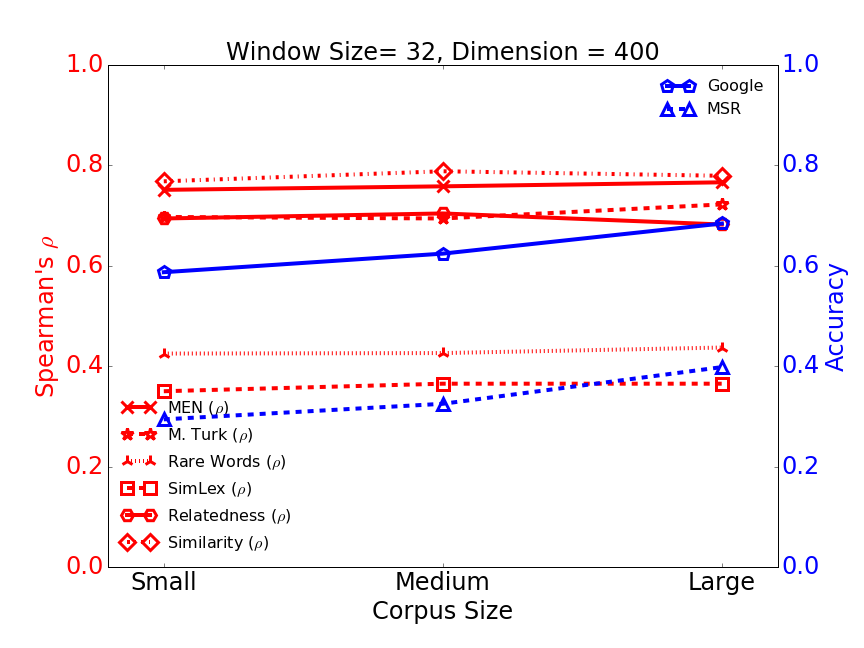}}
        \caption{Task performance with varying corpus sizes.}
		\label{fig:dim_400_corpus}
	\end{minipage}
\end{figure}

The corpus size used to train Swivel was varied (Table~\ref{table:experiment}), the resulting performance on the similarity and analogy tasks are presented in 
Figure~\ref{fig:dim_400_corpus}. 

Interestingly, for the word similarity tasks, models 
trained using a small corpus (with less than 800K tokens) perform almost 
just as well as those trained using larger corpora. The word analogy 
tasks, however, benefit from a larger training corpora. This observation is not 
dissimilar to those found by previous studies 
\cite{Chiu16,FaruquiTRD16}, where different tasks may favour different 
optimal hyper-parameter settings and embedding methodologies.

\section{Discussion}\label{sec:conclusion}
In this paper, we conducted an empirical evaluation of the affect several hyper-parameter settings have on the performance of word embeddings.  
The evaluation was based on standard tasks commonly used in embedding 
evaluation, and all experiments were performed using word embeddings produced by Swivel, a competitive 
embedding model.  The hyper-parameters explored were window 
size, vector dimension size and corpus size, and find interesting observations. 
For similarity tasks, increasing the amount of training 
data has a minimal impact on improving performance.  For other tasks such as the analogy tasks, 
however, the accuracy increases with increased amounts of training data.  In 
general, these hyper-parameters have a window of values that produce 
optimal performance, and that increasing these values beyond the window 
produces little or no performance improvement. These findings will be 
useful in situations where computational resources are limited or dataset size is constrained, 
providing insights for acceptable lower bound values for these 
hyper-parameters.



\bibliography{refs} 
\bibliographystyle{eacl2017}


\appendix
\onecolumn
\section{Supplementary Material}
\label{sec:supplemental}
All the results of the experimental study and evaluation has been shown in Table~\ref{table:results} for further analysis and comparison.
\begin{center}
\begin{longtable}{| p {.8cm} | p {1cm} | p {.8cm} | p {.8cm} | p {1.5cm} | p {1cm} | p {1.2cm} | p {1.8cm} | p {1.5cm} | p {1cm} | p {.8cm}|}
\caption{Results.}
\label{table:results}\\
\hline
\textbf{Data Size} & \textbf{Wind. Size} & \textbf{Dim. Size} & \textbf{MEN} & \textbf{M. Turk} & \textbf{Rare Words} & \textbf{SimLex} & \textbf{Relatedness} & \textbf{Similarity} & \textbf{Google} & \textbf{MSR}\\
\hline
\hline
\endfirsthead
\caption{Results. (continued)}\\
\hline
\textbf{Data Size} & \textbf{Wind. Size} & \textbf{Dim. Size} & \textbf{MEN} & \textbf{M. Turk} & \textbf{Rare Words} & \textbf{SimLex} & \textbf{Relatedness} & \textbf{Similarity} & \textbf{Google} & \textbf{MSR}\\
\hline
\hline
\endhead
\hline
\endfoot
\endlastfoot


S & 1 & 20 & 0.571 & 0.584 & 0.234 & 0.198 & 0.459 & 0.571 & 0.101 & 0.056\\
S & 1 & 100 & 0.663 & 0.597 & 0.303 & 0.282 & 0.617 & 0.695 & 0.366 & 0.218\\
S & 1 & 200 & 0.676 & 0.634 & 0.328 & 0.307 & 0.619 & 0.683 & 0.460 & 0.254\\
S & 1 & 400 & 0.681 & 0.646 & 0.323 & 0.318 & 0.623 & 0.701 & 0.497 & 0.259\\
S & 1 & 500 & 0.692 & 0.608 & 0.331 & 0.330 & 0.627 & 0.696 & 0.487 & 0.256\\
S & 2 & 20 & 0.600 & 0.624 & 0.302 & 0.239 & 0.514 & 0.640 & 0.142 & 0.079\\
S & 2 & 100 & 0.685 & 0.651 & 0.362 & 0.324 & 0.636 & 0.705 & 0.450 & 0.258\\
S & 2 & 200 & 0.710 & 0.652 & 0.365 & 0.330 & 0.657 & 0.743 & 0.517 & 0.284\\
S & 2 & 400 & 0.720 & 0.648 & 0.381 & 0.355 & 0.676 & 0.737 & 0.535 & 0.295\\
S & 2 & 500 & 0.724 & 0.666 & 0.388 & 0.354 & 0.654 & 0.747 & 0.535 & 0.277\\
S & 3 & 20 & 0.614 & 0.638 & 0.307 & 0.240 & 0.510 & 0.671 & 0.161 & 0.072\\
S & 3 & 100 & 0.701 & 0.658 & 0.377 & 0.322 & 0.643 & 0.732 & 0.471 & 0.272\\
S & 3 & 200 & 0.718 & 0.654 & 0.398 & 0.349 & 0.669 & 0.729 & 0.534 & 0.290\\
S & 3 & 400 & 0.732 & 0.679 & 0.398 & 0.349 & 0.671 & 0.749 & 0.550 & 0.293\\
S & 3 & 500 & 0.735 & 0.651 & 0.408 & 0.352 & 0.663 & 0.749 & 0.546 & 0.289\\
S & 4 & 20 & 0.622 & 0.633 & 0.322 & 0.259 & 0.555 & 0.684 & 0.178 & 0.076\\
S & 4 & 100 & 0.715 & 0.655 & 0.386 & 0.306 & 0.661 & 0.744 & 0.475 & 0.262\\
S & 4 & 200 & 0.731 & 0.674 & 0.400 & 0.321 & 0.675 & 0.753 & 0.535 & 0.294\\
S & 4 & 400 & 0.737 & 0.671 & 0.408 & 0.347 & 0.681 & 0.748 & 0.555 & 0.292\\
S & 4 & 500 & 0.737 & 0.676 & 0.414 & 0.353 & 0.675 & 0.757 & 0.555 & 0.287\\
S & 10 & 20 & 0.626 & 0.614 & 0.347 & 0.241 & 0.529 & 0.693 & 0.185 & 0.074\\
S & 10 & 100 & 0.719 & 0.675 & 0.423 & 0.303 & 0.680 & 0.769 & 0.501 & 0.265\\
S & 10 & 200 & 0.737 & 0.702 & 0.414 & 0.329 & 0.706 & 0.772 & 0.553 & 0.296\\
S & 10 & 400 & 0.746 & 0.693 & 0.427 & 0.343 & 0.700 & 0.775 & 0.571 & 0.296\\
S & 10 & 500 & 0.747 & 0.692 & 0.427 & 0.345 & 0.687 & 0.777 & 0.566 & 0.296\\
S & 16 & 20 & 0.632 & 0.619 & 0.373 & 0.248 & 0.503 & 0.671 & 0.184 & 0.070\\
S & 16 & 100 & 0.719 & 0.674 & 0.414 & 0.306 & 0.678 & 0.753 & 0.511 & 0.260\\
S & 16 & 200 & 0.739 & 0.677 & 0.422 & 0.338 & 0.677 & 0.758 & 0.572 & 0.292\\
S & 16 & 400 & 0.748 & 0.688 & 0.433 & 0.346 & 0.705 & 0.784 & 0.581 & 0.299\\
S & 16 & 500 & 0.753 & 0.687 & 0.429 & 0.346 & 0.697 & 0.789 & 0.576 & 0.295\\
S & 32 & 20 & 0.630 & 0.629 & 0.367 & 0.258 & 0.510 & 0.687 & 0.199 & 0.073\\
S & 32 & 100 & 0.723 & 0.667 & 0.417 & 0.316 & 0.670 & 0.752 & 0.521 & 0.263\\
S & 32 & 200 & 0.740 & 0.681 & 0.424 & 0.334 & 0.688 & 0.764 & 0.575 & 0.297\\
S & 32 & 400 & 0.751 & 0.697 & 0.425 & 0.350 & 0.694 & 0.768 & 0.587 & 0.294\\
S & 32 & 500 & 0.753 & 0.695 & 0.425 & 0.363 & 0.697 & 0.766 & 0.587 & 0.297\\
M & 1 & 20 & 0.606 & 0.597 & 0.290 & 0.229 & 0.512 & 0.626 & 0.143 & 0.068\\
M & 1 & 100 & 0.689 & 0.648 & 0.332 & 0.285 & 0.612 & 0.683 & 0.448 & 0.275\\
M & 1 & 200 & 0.706 & 0.658 & 0.360 & 0.315 & 0.599 & 0.696 & 0.522 & 0.311\\
M & 1 & 400 & 0.716 & 0.649 & 0.355 & 0.341 & 0.638 & 0.684 & 0.529 & 0.309\\
M & 1 & 500 & 0.724 & 0.639 & 0.363 & 0.354 & 0.630 & 0.714 & 0.531 & 0.293\\
M & 2 & 20 & 0.635 & 0.627 & 0.343 & 0.256 & 0.534 & 0.672 & 0.175 & 0.069\\
M & 2 & 100 & 0.721 & 0.674 & 0.374 & 0.322 & 0.636 & 0.756 & 0.509 & 0.299\\
M & 2 & 200 & 0.732 & 0.689 & 0.387 & 0.348 & 0.643 & 0.715 & 0.560 & 0.321\\
M & 2 & 400 & 0.743 & 0.678 & 0.383 & 0.370 & 0.668 & 0.737 & 0.574 & 0.330\\
M & 2 & 500 & 0.745 & 0.682 & 0.401 & 0.374 & 0.665 & 0.765 & 0.572 & 0.333\\
M & 3 & 20 & 0.635 & 0.637 & 0.351 & 0.245 & 0.509 & 0.678 & 0.186 & 0.071\\
M & 3 & 100 & 0.719 & 0.681 & 0.385 & 0.333 & 0.677 & 0.759 & 0.533 & 0.302\\
M & 3 & 200 & 0.739 & 0.664 & 0.407 & 0.353 & 0.674 & 0.764 & 0.576 & 0.341\\
M & 3 & 400 & 0.749 & 0.687 & 0.397 & 0.366 & 0.672 & 0.756 & 0.592 & 0.342\\
M & 3 & 500 & 0.751 & 0.659 & 0.412 & 0.380 & 0.686 & 0.782 & 0.587 & 0.338\\
M & 4 & 20 & 0.635 & 0.643 & 0.347 & 0.247 & 0.492 & 0.660 & 0.195 & 0.072\\
M & 4 & 100 & 0.728 & 0.680 & 0.400 & 0.330 & 0.667 & 0.752 & 0.533 & 0.312\\
M & 4 & 200 & 0.743 & 0.694 & 0.399 & 0.357 & 0.684 & 0.756 & 0.590 & 0.336\\
M & 4 & 400 & 0.755 & 0.681 & 0.401 & 0.365 & 0.687 & 0.781 & 0.591 & 0.339\\
M & 4 & 500 & 0.761 & 0.688 & 0.410 & 0.386 & 0.697 & 0.793 & 0.595 & 0.344\\
M & 10 & 20 & 0.632 & 0.636 & 0.361 & 0.254 & 0.513 & 0.681 & 0.200 & 0.060\\
M & 10 & 100 & 0.731 & 0.688 & 0.415 & 0.315 & 0.664 & 0.745 & 0.545 & 0.286\\
M & 10 & 200 & 0.750 & 0.684 & 0.419 & 0.351 & 0.689 & 0.769 & 0.600 & 0.325\\
M & 10 & 400 & 0.758 & 0.697 & 0.417 & 0.369 & 0.692 & 0.784 & 0.610 & 0.337\\
M & 10 & 500 & 0.762 & 0.697 & 0.422 & 0.366 & 0.701 & 0.795 & 0.610 & 0.334\\
M & 16 & 20 & 0.623 & 0.610 & 0.364 & 0.244 & 0.481 & 0.656 & 0.208 & 0.068\\
M & 16 & 100 & 0.729 & 0.687 & 0.419 & 0.318 & 0.682 & 0.773 & 0.548 & 0.289\\
M & 16 & 200 & 0.748 & 0.678 & 0.422 & 0.346 & 0.691 & 0.781 & 0.598 & 0.322\\
M & 16 & 400 & 0.762 & 0.694 & 0.425 & 0.369 & 0.691 & 0.776 & 0.612 & 0.333\\
M & 16 & 500 & 0.763 & 0.704 & 0.429 & 0.363 & 0.699 & 0.791 & 0.619 & 0.324\\
M & 32 & 20 & 0.627 & 0.606 & 0.357 & 0.247 & 0.467 & 0.656 & 0.211 & 0.057\\
M & 32 & 100 & 0.727 & 0.688 & 0.414 & 0.304 & 0.655 & 0.752 & 0.553 & 0.279\\
M & 32 & 200 & 0.745 & 0.688 & 0.419 & 0.348 & 0.684 & 0.775 & 0.608 & 0.326\\
M & 32 & 400 & 0.758 & 0.694 & 0.426 & 0.365 & 0.704 & 0.788 & 0.624 & 0.325\\
M & 32 & 500 & 0.763 & 0.692 & 0.423 & 0.367 & 0.702 & 0.795 & 0.622 & 0.328\\
L & 1 & 20 & 0.620 & 0.631 & 0.292 & 0.222 & 0.490 & 0.640 & 0.214 & 0.072\\
L & 1 & 100 & 0.721 & 0.671 & 0.349 & 0.338 & 0.613 & 0.696 & 0.587 & 0.376\\
L & 1 & 200 & 0.727 & 0.670 & 0.347 & 0.353 & 0.627 & 0.707 & 0.628 & 0.428\\
L & 1 & 400 & 0.752 & 0.692 & 0.374 & 0.376 & 0.639 & 0.712 & 0.641 & 0.443\\
L & 1 & 500 & 0.758 & 0.685 & 0.365 & 0.392 & 0.667 & 0.733 & 0.637 & 0.423\\
L & 2 & 20 & 0.635 & 0.645 & 0.323 & 0.230 & 0.462 & 0.627 & 0.230 & 0.070\\
L & 2 & 100 & 0.736 & 0.693 & 0.390 & 0.331 & 0.631 & 0.746 & 0.605 & 0.384\\
L & 2 & 200 & 0.747 & 0.692 & 0.395 & 0.365 & 0.634 & 0.727 & 0.665 & 0.432\\
L & 2 & 400 & 0.759 & 0.701 & 0.403 & 0.389 & 0.656 & 0.735 & 0.670 & 0.446\\
L & 2 & 500 & 0.770 & 0.710 & 0.416 & 0.405 & 0.686 & 0.757 & 0.669 & 0.448\\
L & 3 & 20 & 0.626 & 0.634 & 0.340 & 0.231 & 0.462 & 0.664 & 0.228 & 0.063\\
L & 3 & 100 & 0.740 & 0.701 & 0.407 & 0.338 & 0.656 & 0.761 & 0.604 & 0.368\\
L & 3 & 200 & 0.753 & 0.717 & 0.410 & 0.359 & 0.648 & 0.758 & 0.659 & 0.426\\
L & 3 & 400 & 0.761 & 0.711 & 0.414 & 0.382 & 0.661 & 0.748 & 0.682 & 0.444\\
L & 3 & 500 & 0.771 & 0.707 & 0.422 & 0.394 & 0.676 & 0.781 & 0.672 & 0.439\\
L & 4 & 20 & 0.631 & 0.628 & 0.341 & 0.229 & 0.442 & 0.636 & 0.228 & 0.054\\
L & 4 & 100 & 0.739 & 0.711 & 0.398 & 0.328 & 0.641 & 0.760 & 0.606 & 0.363\\
L & 4 & 200 & 0.754 & 0.723 & 0.416 & 0.369 & 0.667 & 0.769 & 0.656 & 0.420\\
L & 4 & 400 & 0.763 & 0.715 & 0.425 & 0.380 & 0.654 & 0.773 & 0.672 & 0.435\\
L & 4 & 500 & 0.773 & 0.721 & 0.435 & 0.386 & 0.694 & 0.789 & 0.673 & 0.427\\
L & 10 & 20 & 0.627 & 0.618 & 0.350 & 0.215 & 0.447 & 0.635 & 0.213 & 0.047\\
L & 10 & 100 & 0.735 & 0.712 & 0.409 & 0.313 & 0.654 & 0.768 & 0.580 & 0.317\\
L & 10 & 200 & 0.756 & 0.710 & 0.423 & 0.348 & 0.672 & 0.779 & 0.657 & 0.390\\
L & 10 & 400 & 0.766 & 0.718 & 0.432 & 0.376 & 0.682 & 0.787 & 0.682 & 0.415\\
L & 10 & 500 & 0.771 & 0.725 & 0.439 & 0.377 & 0.685 & 0.784 & 0.684 & 0.422\\
L & 16 & 20 & 0.620 & 0.615 & 0.356 & 0.220 & 0.428 & 0.610 & 0.214 & 0.047\\
L & 16 & 100 & 0.735 & 0.709 & 0.412 & 0.306 & 0.644 & 0.758 & 0.570 & 0.305\\
L & 16 & 200 & 0.759 & 0.716 & 0.432 & 0.341 & 0.683 & 0.788 & 0.658 & 0.377\\
L & 16 & 400 & 0.765 & 0.726 & 0.436 & 0.373 & 0.675 & 0.775 & 0.681 & 0.409\\
L & 16 & 500 & 0.771 & 0.724 & 0.438 & 0.372 & 0.683 & 0.779 & 0.691 & 0.414\\
L & 32 & 20 & 0.614 & 0.626 & 0.353 & 0.211 & 0.397 & 0.607 & 0.213 & 0.046\\
L & 32 & 100 & 0.730 & 0.695 & 0.410 & 0.296 & 0.634 & 0.746 & 0.555 & 0.283\\
L & 32 & 200 & 0.753 & 0.713 & 0.425 & 0.337 & 0.662 & 0.774 & 0.647 & 0.358\\
L & 32 & 400 & 0.766 & 0.722 & 0.437 & 0.365 & 0.682 & 0.779 & 0.684 & 0.398\\
L & 32 & 500 & 0.767 & 0.716 & 0.440 & 0.368 & 0.688 & 0.777&0.684&0.404\\
\hline
\end{longtable}
\end{center}

\end{document}